\documentclass[twoside]{article}

\usepackage[accepted]{aistats2024}
\usepackage{amsfonts}
\usepackage{hyperref}
\usepackage{graphicx}
\usepackage{bbm}
\usepackage{amsmath}
\usepackage{amsthm}
\usepackage{xcolor}

\newtheorem{theorem}{Theorem}
\newtheorem{lemma}{Lemma}
\newtheorem{definition}{Definition}

\newtheorem{corollary}{Corollary}

\newcommand{\x}{\mathbf{x}}

\newcommand{\ra}{\mathbf{u}}

\newcommand{\z}{\mathbf{z}}
\newcommand{\Br}{\mathcal{B}_r}
\newcommand{\E}{\mathbb{E}}
\newcommand{\Sj}{\mathcal{S}_j}
\newcommand{\Emu}{\mathbb{E}_{\x\sim\mu}}

\newcommand{\EBr}{\underset{\z\in\Br}{\E}}

\newcommand{\uu}{\mathbf{u}}

\newcommand{\AS}{\mathbf{C}_{f,\mu}}
\newcommand{\ASE}{\mathbf{B}_{f,\mu}}

\begin{document}



\twocolumn[

\aistatstitle{Surrogate Active Subspaces for Jump-Discontinuous Functions}

\aistatsauthor{Nathan Wycoff}

\aistatsaddress{The McCourt School's Massive Data Institute, Georgetown University, Washington, D.C.} ]

\begin{abstract}
%\todo{New Abstract}
%Surrogate models in the form of Gaussian processes have found great success in applications to emulation of computational models, executing tasks such as optimization, calibration, and sensitivity analysis.
%Standard surrogate models are continuous functions, and so are not necessarily applicable to computational models which exhibit discontinuity, prompting much research into similarly discontinuous surrogates.
%In this article, we show that a standard continuous surrogate model can yield an informative gradient-based sensitivity analysis when deployed on a piecewise constant function.
%We numerically study the Gibbs phenomena associated with a Gaussian process fit to a discontinous function, as well as the analytic behavior of simple surrogate estimates of the active subspace in such contexts.
%These form the basis of a methodology which extends gradient-based sensitivity analysis to piecewise constant functions by applying the sensitivity to kernel approximations of those functions.
%We validate our methodology on toy functions, comparing it to other sensitivity metrics.
%Finally, we deploy our methodology on *Flee*, an agent-based model of refugee movement, yielding novel insights into which parameters of the simulation are most important across 8 refugee crises in Africa and the Middle East. 
Surrogate modeling and active subspaces have emerged as powerful paradigms in computational science and engineering. 
Porting such techniques to computational models in the social sciences brings into sharp relief their limitations in dealing with discontinuous simulators, such as Agent-Based Models, which have discrete outputs.
Nevertheless, prior applied work has shown that surrogate estimates of active subspaces for such estimators can yield interesting results. 
But given that active subspaces are defined by way of gradients, it is not clear what quantity is being estimated when this methodology is applied to a discontinuous simulator.
We begin this article by showing some pathologies that can arise when conducting such an analysis.
This motivates an extension of active subspaces to discontinuous functions, clarifying what is actually being estimated in such analyses.
We also conduct numerical experiments on synthetic test functions to compare Gaussian process estimates of active subspaces on continuous and discontinuous functions.
Finally, we deploy our methodology on Flee, an agent-based model of refugee movement, yielding novel insights into which parameters of the simulation are most important across 8 displacement crises in Africa and the Middle East. 
\end{abstract}

\section{Introduction}

In most fields of science and engineering, cutting edge mathematical models of phenomena are not susceptible to closed-form mathematical analysis, and must instead be studied via numerical simulation.
One approach to conducting such a study is via a \textit{sensitivity analysis}, which aims to determine what input parameters, or combinations thereof, the output of a simulator is most influenced by.
In this article, we will be concerned with the Active Subspace Method \cite{constantine2015active} (ASM, see Section \ref{sec:bg_asm}), a form of global sensitivity analysis based on analyzing the gradient of the target function.
When the computer simulation is computationally expensive, a common approach to studying it is to fit a \textit{surrogate model}, that is, to estimate a flexible statistical model to sampled input-output pairs. 
Furthermore, some early work has explored porting these tools to simulators of social scientific phenomena, such as Agent-Based Models (ABMs).
But ABMs represent the sum of discrete choices made by individuals, and so are inherently discontinuous. 
%Fitting a well-behaved surrogate to simulators with numerical noise or other pathologies can also be a benefit in and of itself.
Nevertheless, nothing prevents an analyst from fitting a smooth surrogate model to a discontinuous simulator and calculating the surrogate's active subspace.
For instance, \cite{notestine2022sensitivity} computes various surrogate estimates of the active subspace of an ABM of social unrest, and shows that useful conclusions can be drawn from such an analysis.
We also find promising results in our case study on an ABM of forced displacement, where a surrogate active subspace analysis offers novel conclusions and improves predictive accuracy.
But it's not clear what is actually being estimated, since an ABM is not differentiable and is almost everywhere constant, so the ``true" active subspace is undefined or $\mathbf{0}$. 
We might hope that another sensitivity analytic framework, like Sufficient Dimension Reduction (see Section \ref{sec:bg_sdr}), can tell us what's going on. 
But our Corollary \ref{cor:nosdr} shows that this is not the case.
In this article, we develop an extension of active subspaces to certain functions with jump-discontinuities.
We find that asymptotically, a surrogate active subspace analysis will favor discontinuous directions of variation over continuous ones, and numerically find that in finite samples, the sample size implicitly parameterizes a trade-off between continuous and discontinuous directions of variation.

Though there seems to be significant demand for surrogate modeling of discontinuous simulators as evidenced by the plethora of applied articles expounding their usefulness (see Section \ref{sec:bg_disc}), the surrogate methodologist's conception of a ``black-box" is overwhelmingly a continuous one. 
This article aims to play some small part in filling this methodological gap by making the following contributions:
\begin{enumerate}
    \item In Section \ref{sec:motivation}, we show that the use of surrogate active subspaces on simulators with jump discontinuities can lead to unexpected pathologies.
    \item We develop an extension of active subspaces to discontinuous simulators to explain the observed pathologies and provide a theoretical basis for surrogate active subspace analysis of discontinuous simulators in Section \ref{sec:theory}.
    \item Section \ref{sec:kern} studies the fitness of various Gaussian process kernels for estimating the active subspace of discontinuous functions, finding rougher ones to be best.
    \item In our case study of Section \ref{sec:flee}, we show that surrogate active subspace analysis can lead to quantitative results superior to dimension reduction which avoids gradients altogether and to meaningful qualitative insights.
\end{enumerate}
Additionally to our main contributions, the pathologies we reveal in surrogate discontinuous active subspace analysis open the door for significant future work which we briefly overview in Section \ref{sec:disc}.

\section{Motivation}\label{sec:motivation}

We begin this section with an overview of the \textit{Flee} ABM, the application that motivated this research.
Subsequently, we present two distressing observations about the empirical behavior of surrogate sensitivity analysis on discontinuous functions.

\subsection{The \textit{Flee} Simulator}\label{sec:mot_flee}

The $21$st century will experience unprecedented migration due to the changing climate, both environmental (\cite{wrathall2019meeting}) and political (including the ongoing mass-displacement events in Ukraine and Gaza), which motivates the study of human migration.
The Flee simulator\footnote{Flee is licensed under BSD-3.} \cite{suleimenova2017generalized} is an ABM of the journeys of forcibly displaced persons.
Agents are displaced over time and move from populated areas to refugee camps and neighboring countries according to simulation parameters, of which we study 7 in this article (see Table \ref{table:params}).
Beyond these parameters, \textit{Flee} also requires a spatial context in which to simulate movement.
We study six crises resulting in forced displacement provided by \textit{Flee}, namely the South Sudanese civil war in 2014, the civil war in the Central African Republic in 2013, the 2012 Malian \textit{coup d'\'etat}, the 2013 escalation of Syria's civil war, the Ethiopian civil war of 2020 and the 2015 civil unrest in Burundi. 
In each case, the simulator compares its estimates of displacement with ground truth data and provides a scalar error estimate.
We wish to study the sensitivity of this error with respect to the model parameters for each context individually.

% latex table generated in R 4.3.1 by xtable 1.8-4 package
% Sun Oct 15 21:43:07 2023
\begin{table}[ht]
\centering
\begin{tabular}{|rrrr|}
  \hline
 Parameter & Min & Max & Default \\ 
  \hline
max\_move\_speed & 0.0 & 40000 & 200 \\ 
  max\_walk\_speed & 0.0 & 40000 & 35 \\ 
  camp\_move\_chance & 0.0 & 1.0 & 0.0 \\ 
  conflict\_move\_chance & 0.0 & 1.0 & 1.0 \\ 
  default\_move\_chance & 0.0 & 1.0 & 0.3 \\ 
  camp\_weight & 1.0 & 10.0 & 2.0 \\ 
  conflict\_weight & 0.1 & 1.0 & 0.2 \\ 
   \hline
\end{tabular}
\caption{\textit{Flee} model parameters.}
\label{table:params}
\end{table}

\subsection{Divergence of the Classical Active Subspace Estimate}

\begin{figure}
    \centering
    
    \includegraphics[width=0.4\textwidth]{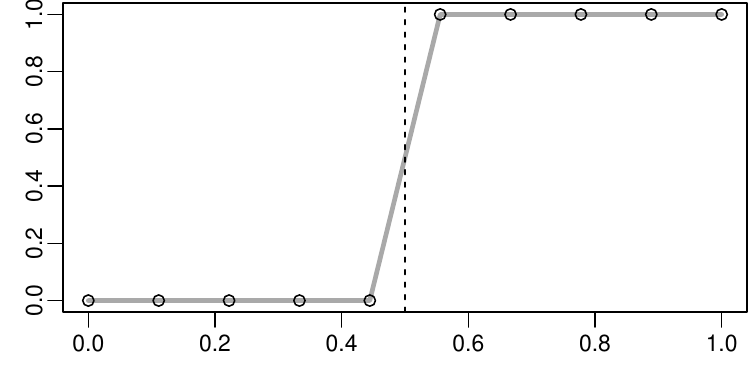}
    
    \includegraphics[width=0.4\textwidth]{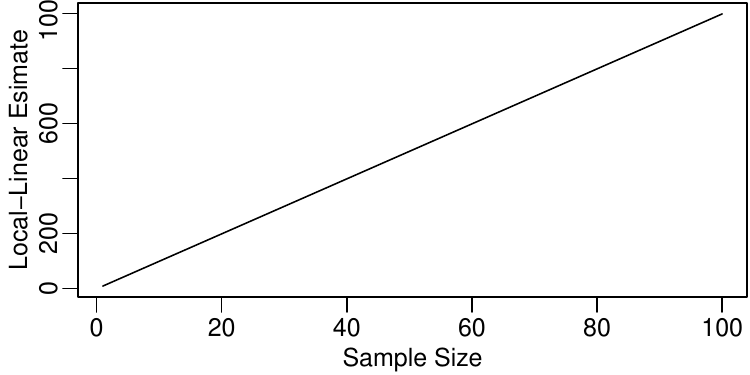}
    \caption{\textit{Top:} Piecewise linear interpolant of the heaviside step function.
    \textit{Bottom:} The piecewise linear estimate of active subspace, which diverges.}
    \label{fig:1d_disc}
\end{figure}

Though the notion of an active subspace is not well defined for functions which are not differentiable such as ABMs, we might hope that fitting an almost-everywhere-continuous surrogate would lead to a reasonable estimate. 
%In this section, we investigate a simple 1 dimensional function which shows that there are ontological issues associated with a surrogate active subspace analysis of a discontinuous function, even in one dimension.
%We consider heaviside step function centered at $0.5$ to agree with the standard setup in computer experiments.
In this section, we consider a 1 dimensional test function given by the heaviside step function centered at $0.5$.
We interpolate this function at an evenly spaced grid of an $n_g=2k$ points by simply drawing a line between subsequent observations (Figure \ref{fig:1d_disc}, top).
All line segments are of slope zero except for the center-most segment, which has slope $n-1$, and extent $\frac{1}{n-1}$. 
Hence, if $s_i$ gives the slope of the $i$th line segment, the active subspace of the surrogate is given by (see Section \ref{sec:bg_asm}) $\frac{1}{n_g-1}\sum_{i=1}^{n_g-1} s_i^2 = \frac{1}{n_g-1}(n_g-1)^2 = n_g-1$.
We thus see that the active subspace estimate diverges as $n_g\to\infty$ (Figure \ref{fig:1d_disc}, bottom). 
%The purpose of the next section will be to develop a generalized definition of the active subspace which converges when suitably renormalized.
It is simple in this case to normalize by $n$ to avoid divergence, and in any case, the scaling of the sensitivity analysis is not important.
However, this divergence is indicative of a deeper issue which can lead to unexpected outcomes, as we discuss next.

\subsection{Contradictory Sensitivity Analyses in a Mixed Simulator}\label{sec:mot_mix}

\begin{figure}
    \centering
    \includegraphics[width=0.50\linewidth,trim=0 0 5em 0,clip]{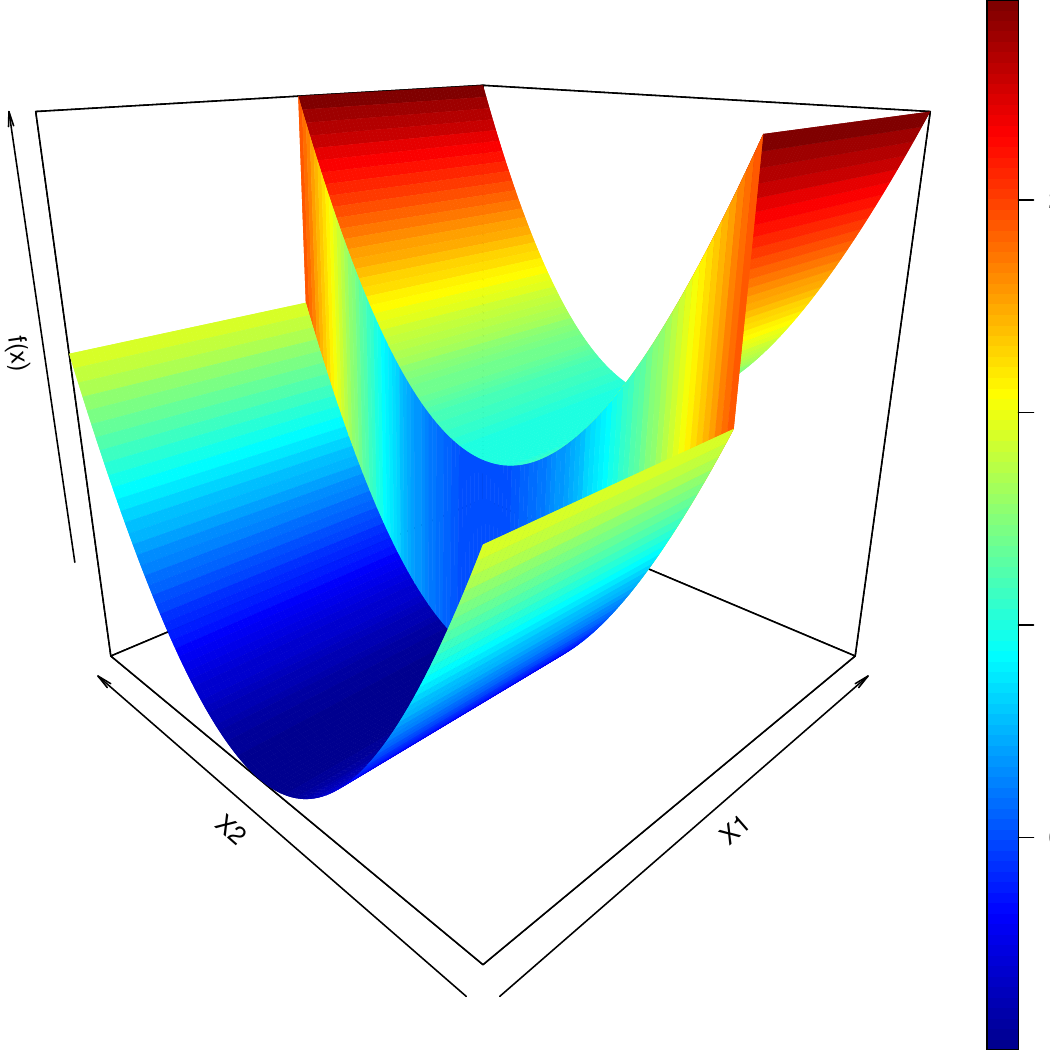}

    \includegraphics[width=0.4\textwidth]{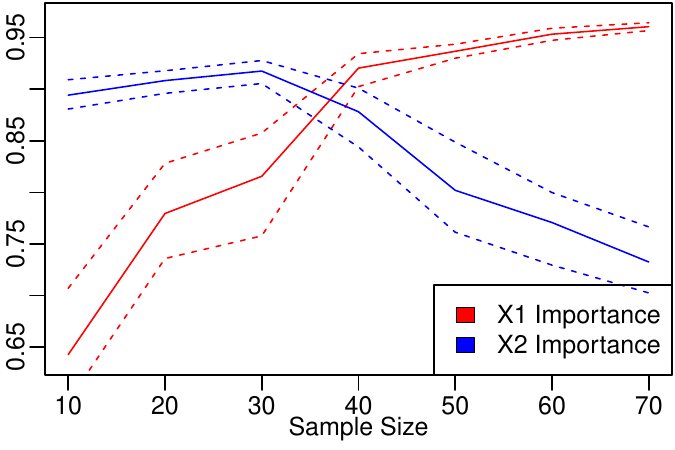}
    \caption{\textit{Top:} A function with a discontinuity along the $x_1$ direction and a smooth quadratic form along the $x_2$ direction. \textit{Bottom:} As the sample size increases, the expected importance of the discontinuous direction overtakes the continuous one.}
    \label{fig:disc_wins}
\end{figure}

We now consider the two dimensional function $f(\x) = \mathbbm{1}_{[x_1\geq 0.5]} + 6 (x_2-0.5)^2$, which varies smoothly along $x_2$ but has a jump along $x_1$ (see Figure \ref{fig:disc_wins}, top).
We draw $N$ random points in the unit square and use these to compute a Gaussian process surrogate estimate of the diagonal elements of the active subspace matrix normalized to have norm 1, which are indicators of variable importance (see Section \ref{sec:bg_disc}). 
For $N\leq30$, we see that the analysis consistently reports that $x_2$, the smooth variable is more important (Figure \ref{fig:disc_wins}, bottom) than $x_1$.
However, for $N\geq 40$, this is reversed.
As the design points are placed closer together, the sensitivity estimate in the smooth direction stabilizes, while that in the discontinuous direction diverges.
Section \ref{sec:theory} develops theory explaining this phenomenon, but we first catch up on the needed methodological background.

\section{Background}\label{sec:bg}

We review surrogate modeling of discontinuous simulators and some concepts from linear sensitivity analysis.

\subsection{Surrogate Modeling of Computer Experiments}
The practice of \textit{Surrogate Modeling} \cite{gramacy2020surrogates} corresponds to the use of flexible statistical models to approximate parameterized computer simulations, conceptualized as input-output maps.
In this article, we will be interested in studying surrogates of a black-box function $f$ mapping $\mathcal{X}\subseteq\mathbb{R}^P\to\mathbb{R}$.
Some of our technical results rely on $\mathcal{X}$ being compact, and in the numerical studies it will be the unit hypercube $[0,1]^P$.

\subsection{Surrogacy for Discontinuous Simulators}\label{sec:bg_disc}

Sometimes, there's an important and well understood discontinuity that we'd like our surrogate model to preserve. 
For instance, in aerodynamics, the trans-sonic barrier leads to completely different dynamics on one side from the other.
\cite{dupuis2018surrogate} use use different surrogates for subsonic and various supersonic operating conditions to accurately capture the jump as well as faithfully approximate the truth on either side.
In the case of ABMs, this is due to their being a model of a sum of discrete choices.
In other circumstances, the discontinuity can be a nuisance caused not by an underlying natural phenomenon of interest but rather due to numerical noise or nonconvergence of the simulation.
In such cases, the hope is rather that the surrogate will paper over the inadequacies of the model.
Take \cite{huang2020site}, whose simulator exhibited discontinuous jumps related to tolerance parameters. 
Between these two extremes lie a number of other possible situations.
In the domain of Structural Optimization for Crashworthiness, the simulator studied by \cite{niutta2018surrogate} has important discontinuities, but the nature and number of them is not known \textit{a priori}.
The authors estimate the number and location of discontinuities using a combined surrogate approach.
\cite{gorodetsky2014efficient} propose methodology for estimating the location at which a jump occurs by examining a polynomial interpolant of the function.
\cite{audet2022escaping} propose an approach for Black-Box optimization under the constraint that the optimum cannot lie near the unknown locations of discontinuity.

We see that some authors choose to model piecewise-discontinuous functions with similarly piecewise-discontinuous surrogates, while others use a global smooth surrogate.
In this article, we will study the latter approach, finding both positive and negative results.
Some pathologies of fitting continuous interpolants to discontinuous functions have been long known, such as the tendency for Fourier approximations to oscillate when approximating discontinuous functions, which is known as the Gibbs Phenomenon \cite[Chapter 14.5]{arfken1972mathematical}.
Indeed, this is true of any global smooth approximant to a discontinuous function \cite{butzer1987approximation}.

\subsection{Gradient-Based Global Sensitivity Analysis}\label{sec:bg_asm}

For a smooth function $f$, the gradient $\nabla f(\x)$ is a natural way of quantifying the sensitivity of an output to an input. 
One strategy for turning this local estimate of sensitivity into a global one is to integrate it over the parameter space: $\AS=\int_{\mathcal{X}}\nabla f(\x) \nabla f(\x)^\top d\mu(x)$.
Here, $\mu$ is a probability measure on the input space.
In the computational engineering literature, this often goes by the name Active Subspace Method (ASM) \cite{constantine2015active}, and such expected gradient outer products have also been called Average Derivative Functionals in the observational context \cite{samarov1993exploring}.
Examination of the eigendecomposition of $\AS$ gives important linear combinations of inputs.
When the gradient of the target function is available, a Monte-Carlo estimate is straightforward to form \cite{constantine2014computing}. 
Otherwise, the strategy of computing the active subspace of a surrogate model can be deployed \cite{palar2018accuracy}.
In the observational context, \cite{fukumizu2014gradient} produce a kernel estimate of the active subspace with respect to the empirical measure of a given sample.
\cite{wycoff2021sequential} showed that if the surrogate model is a Gaussian process with certain kernel functions, the active subspace is available in closed form.

\subsection{Sufficient Dimension Reduction }\label{sec:bg_sdr}

Another perspective on linear dimension reduction is that of Sufficient Dimension Reduction (SDR). 
Given a measure $\mu$ on $\x$, we say that $\mathcal{U}$ is a sufficient reduction if $P(y|\x)=P(y|\mathbf{U}\x)$ \cite{cook1994interpretation}, where $\mathbf{U}$ is a matrix with range $\mathcal{U}$ (though there are also slightly different definitions and related concepts \cite{adragni2009sufficient}).
See \cite{ma2013review} for a review.
One approach to estimating sufficient reductions is Sliced Inverse Regression \cite{li1991sliced}, which splits the response $y$ into bins before taking the mean $\x$ value in each bin and performing PCA on the resulting matrix.

\subsection{Ridge Functions}
Related to both active subspaces and SDR is the concept of a \textit{ridge function} \cite{logan1975optimal}, that is, a function $f:\mathbb{R}^P\to\mathbb{R}$ which takes $\x\to g(\mathbf{A}\x)$ with $\mathbf{A}\in\mathbb{R}^{R\times P}$ for $R<P$ and $g:\mathbb{R}^R\to\mathbb{R}$.
Like SDR, it encapsulates the idea of relationships which depend \textit{soley} on certain input dimensions.
By comparison, the concept of ASM is fuzzier, allowing \textit{some} variation in all directions, but focusing it along certain ones.

\section{An Extension of Active Subspaces to Discontinuous Functions}\label{sec:theory}

In this section, we'll develop an extension of Active Subspaces to discontinuous functions in order to explain what is being estimated by the active subspace of a surrogate model fit to a discontinuous simulator.
The proofs of all results are given in the Supplementary Materials.
Throughout this section, $\Vert.\Vert$ will refer to the Euclidean norm, $C^1$ represents the function space of functions once differentiable on $\mathcal{X}$, $\Br^P$ is the ball of radius $r$ in Euclidean $P$-space, $\Gamma(x)$ refers to the special function, and $\mu$ is a probability measure on $\mathcal{X}$ and for some results it will be assumed to have Lebesgue density $\delta$.
We will consider simulators abstracted mathematically as functions given by the sum of characteristic functions for sets parameterized by a differentiable function together with a smooth term, that is, 
$f(\x) = \sum_{j=1}^J c_j \mathbbm{1}_{[\x\in\Sj]} + g(\x)$ where $\mathbbm{1}_{[\x\in\mathcal{A}]}$ is the funtion taking value $1$ if $\x\in\mathcal{A}$ and zero otherwise, 
$\Sj = \{\x\in\mathcal{X} : h_j(\x) \leq 0\}$
\footnote{Our results hold for $\Sj$ defined either by strict or nonstrict inequality, leading to either open or closed sets.
For notational simplicity, we use closed sets throughout.}
where $h_j\in C^1$ for all $j$,
and $g\in C^1$.

Our extension will be built on a continuous analog to a regression coefficient, intuitively given by the limit of the OLS estimate given by sampling points uniformly within a radius $r$ of a given point $\x$ as the sample size tends to infinity.
\begin{definition}
    $\beta_r(\x) = \EBr[\z\z^\top]^{-1}\EBr[\z y(\x+\z)]$.
\end{definition}
%\todo{We don't really use the expectation form anywhere; would it have been simpler if we did?}
%Intuitively, this is the local OLS estimate of the function near $\x_g$ with an infinitely dense grid.
%Though $\beta_r(\x)$ is not computable, we can get approximations by approximating the integrals. 

To work with $\beta^r(\x)$, we will need the following elementary results, where the Gamma function arises from the volume of the $P$-ball:

\begin{enumerate}
    \item $\int_{\z\in\Br^P} z_i^2d\z = \frac{\pi^{\frac{P}{2}}r^{P+2}}{2\Gamma(\frac{P+4}{2})} := \xi_P r^{P+2}$
    \item $\mathbb{E}_{\z\in\Br^P} [z_i^2] = \frac{r^2}{P+2}$
\end{enumerate}

%\textbf{Mini Lemma 1:}
%
%\begin{equation}
%    \int_{\x\in\Br^P} x_i^2d\x = \frac{\pi^{\frac{P}{2}}r^{P+2}}{2\Gamma(\frac{P+4}{2})} = \xi_P r^{P+2}
%\end{equation}
%
%\textbf{Mini Lemma 2:}
%
%\begin{equation}
%    \mathbb{E}_{\x\in\Br^P} [x_i^2] = \frac{r^2}{P+2}
%\end{equation}

%We have that $\EBr[\x\x^\top]^{-1} = \frac{P+2}{r^2}\mathbf{I}$ (Mini Lemma 2).

This leads to the following result.

\begin{lemma}\label{lem:beta_grad}
    If $f$ consists only of a smooth term $g$ we have that $\underset{r\to 0}{\lim} \beta_r(\x) = \nabla f (\x)$. 
\end{lemma}

Lemma \ref{lem:beta_grad} tells us that $\beta^r(\x)$ may be viewed as an extension of the gradient to possibly discontinuous functions. 
This motivates the following natural definition for an extension of the active subspace:

\begin{definition}
    Let $\ASE^r = \Emu[\beta_r(\x)\beta_r(\x)^\top]$. We define our active subspace extension $\ASE = \underset{r\to 0}{\lim} A_P r \ASE^r$, where $A_P$ is a constant depending only on $P$ given in the Supplementary Material.
\end{definition}

The above definition together with Lemma \ref{lem:beta_grad} yields the following theorem, which compares our extension to the original active subspace on differentiable functions.

\begin{theorem}\label{thm:ext}
    If $f$ is once differentiable (i.e. $f=g$) and bounded on $\mathcal{X}$, compact, then $\underset{r\to 0}{\lim}\ASE^r = \AS$.
\end{theorem}

This theorem confirms the status of $\ASE$ as an extension of $\AS$, insofar as it agrees with it for sufficiently regular functions.
Next we investigate some properties of $\ASE$ which apply in the general case where $f$ is discontinuous.

%\begin{remark}\label{rem:ext}
%    \todo{This might not actually be necessary in this latest formulation.}
%    Note that we have required $f \in C^2$ when we show that our extension agrees with the original active subspace definition.
%    But the original active subspace is defined on $C^1$, and though our extension is defined on this space, it need not agree with the original active subspace. 
%    Therefore, our extensions should be viewed as an extension of the active subspace as an object defined on $C^2$ to the function space $\Fs$, rather than an extension of the notion of an active subspace \textit{tout court}.
%\end{remark}

\begin{lemma}
    If $f(\x)$ is constant along dimension $\uu$, and $\mu$ is translation-invariant along $\uu$, then $\uu^\top\beta^r(\x) = 0.$
\end{lemma}

This lemma shows us that even for finite $r$, the gradient analogue $\beta^r(\x)$ will always point in directions which the target function vary in.
It leads to the below theorem.

\begin{theorem}
    If $f(\x) = g(\mathbf{A} \x)$ with $\mathbf{A}\in\mathbb{R}^{R\times P}$ and $g: \mathbb{R}^R\to\mathbb{R}$, $\textrm{Range}(\ASE^r) \subseteq \textrm{Range}(\mathbf{A})$.
\end{theorem}

This theorem represents our first positive result, showing that ridge functions, including discontinuous ones, have their ridge structure respected by the extended active subspace.
But the active subspace on continuous functions can tell us more than simply whether or not a given function has ridge structure or not; it also gives us the relative importance of different directions defined in a sum of squares sense. 
We now turn to investigating analogous properties of our proposed extension, starting in one dimension to build intuition. 

\begin{lemma}
    If $f:\mathbb{R}\to\mathbb{R}$ is a linear combination of translated heaviside functions, that is $f(x) = \sum_{j=1}^Jc_j\mathbbm{1}_{[x\leq\tau_j]}$, and $\mu$ is Lebesgue-continuous with differentiable density $\delta$, then $\underset{r\to 0}{\lim} \frac{5r}{3}\ASE^r = \sum_{j=1}^J c_j^2 \delta(\tau_j)$.
\end{lemma}

This lemma shows us that, when properly normalized, the active subspace extension gives the sum of squared jumps of a discontinuous function, weighted by the density of the measure with respect to which it is defined. 
The following theorem extends this understanding to $P$ dimensions.

\begin{theorem}
    Given the constant $A_P$ depending on $P$, compact $\mathcal{X}$ and assuming that the discontinuities do not overlap substantially as made precise in the Supplementary Material, we have that $\underset{r\to 0}{\lim} A_P r \ASE^r = \sum_{j=1}^J c_j^2 \big[\int_{\{\x : h_j(\x)=0\}} \frac{1}{\Vert\nabla h_j(\x)\Vert_2^2} \nabla h_j(\x) \nabla h_j(\x)^\top \delta(\x)d\x\big]$.
\end{theorem}

Intuitively, the active subspace extension is given by a weighted sum of an active subspace analogue of the functions parameterizing the jump points, weighted by the squared size of the jump and with a degenerate measure confined to the null-set of $h_j$.
However, unlike the standard active subspace definition, note that the expression
$\frac{1}{\Vert\nabla h_j(\x)\Vert_2^2} \nabla h_j(\x) \nabla h_j(\x)^\top $
is invariant to smooth monotonic transformation to any $h_j$, which is necessary given that this kind of transformation will have no effect on $f$.
%Note also that the expression in the preceding theorem does not depend on g. 
The following is an immediate consequence of the fact that the expression in the preceding theorem does not depend on $g$, and is our main negative result.

\begin{corollary}
    For $f$ with both smooth and discontinuous components, the range of $\ASE$ does not necessarily contain the sufficient directions. \label{cor:nosdr}
\end{corollary}

This tells us that the extended active subspace ignores smooth directions of hybrid smooth-discontinuous functions.
It helps to explain the contradictory behavior we observed when estimating a surrogate's active subspace fit to a discontinuous function in Section \ref{sec:mot_mix}.
    
\section{Numerical Study of Kernel Estimates}\label{sec:kern}

\begin{figure*}
    \centering

    \includegraphics[width=0.23\textwidth]{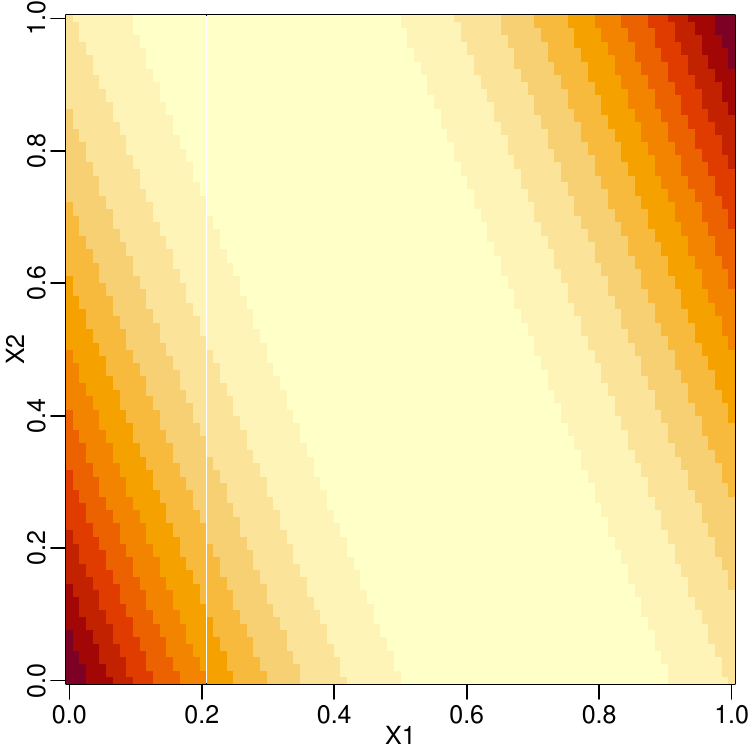}
    \includegraphics[width=0.23\textwidth]{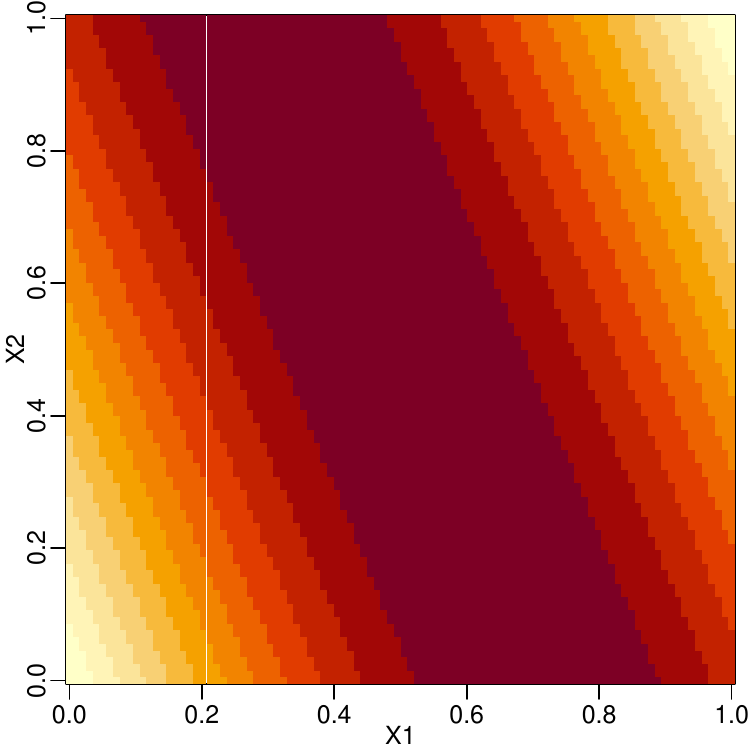}
    \includegraphics[width=0.23\textwidth]{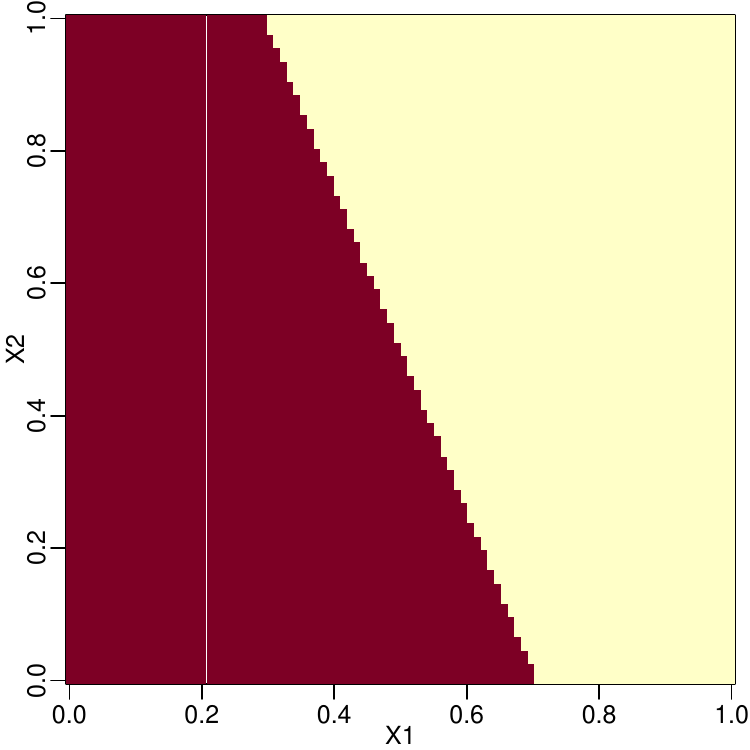}
    \includegraphics[width=0.23\textwidth]{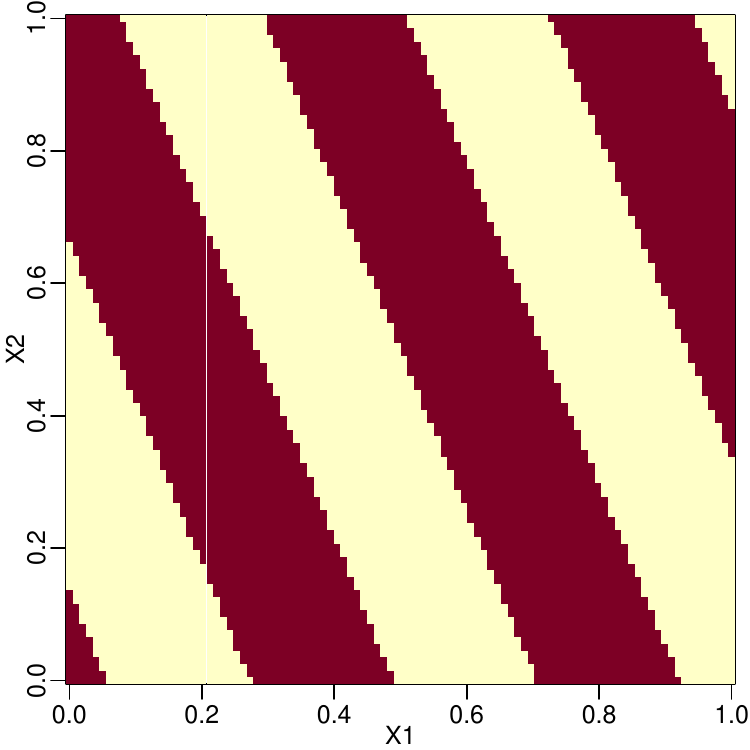}
    
    \includegraphics[width=0.23\textwidth]{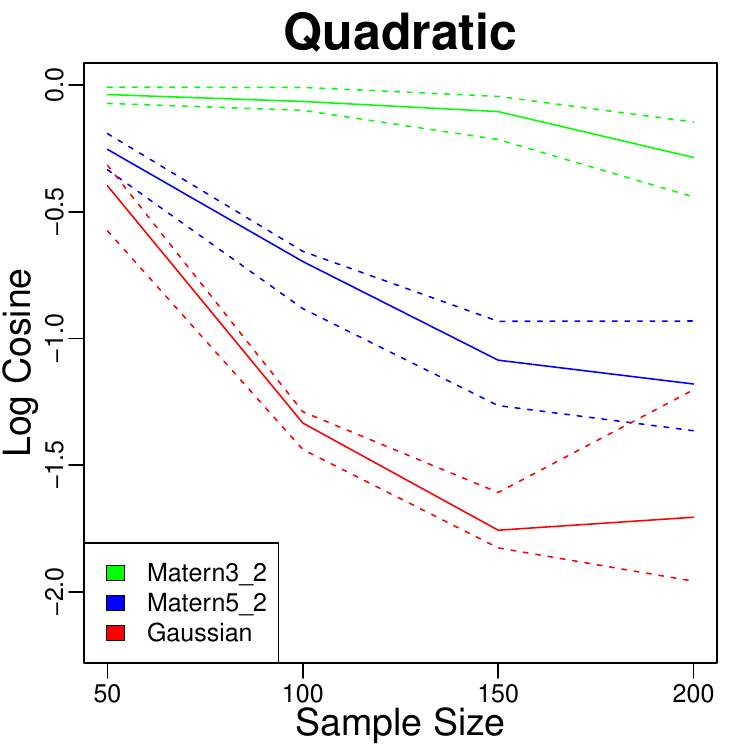}
    \includegraphics[width=0.23\textwidth]{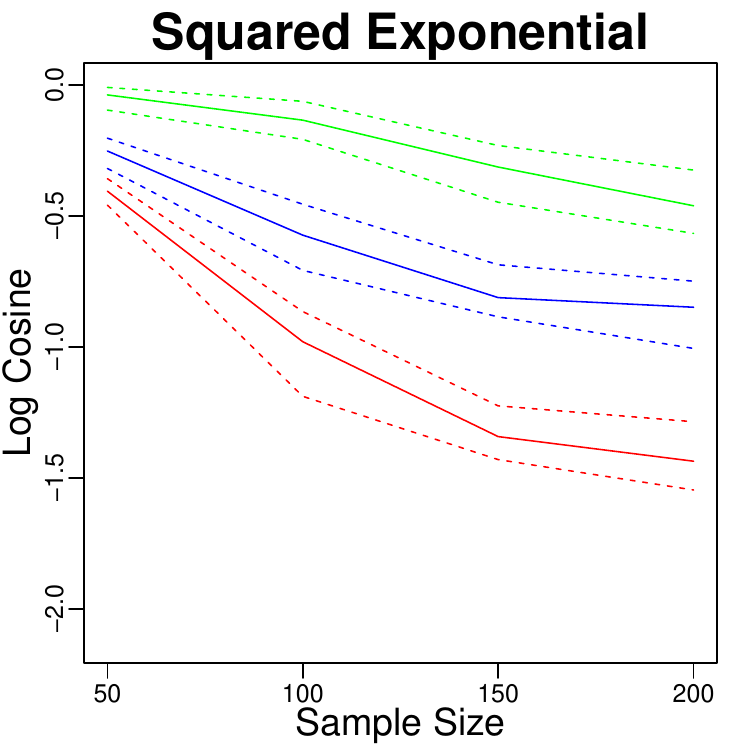}
    \includegraphics[width=0.23\textwidth]{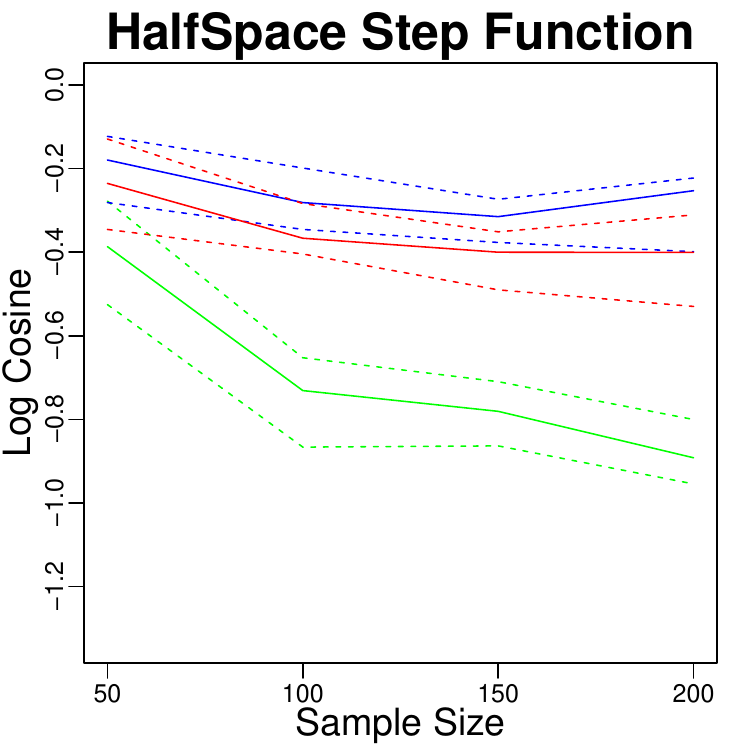}
    \includegraphics[width=0.23\textwidth]{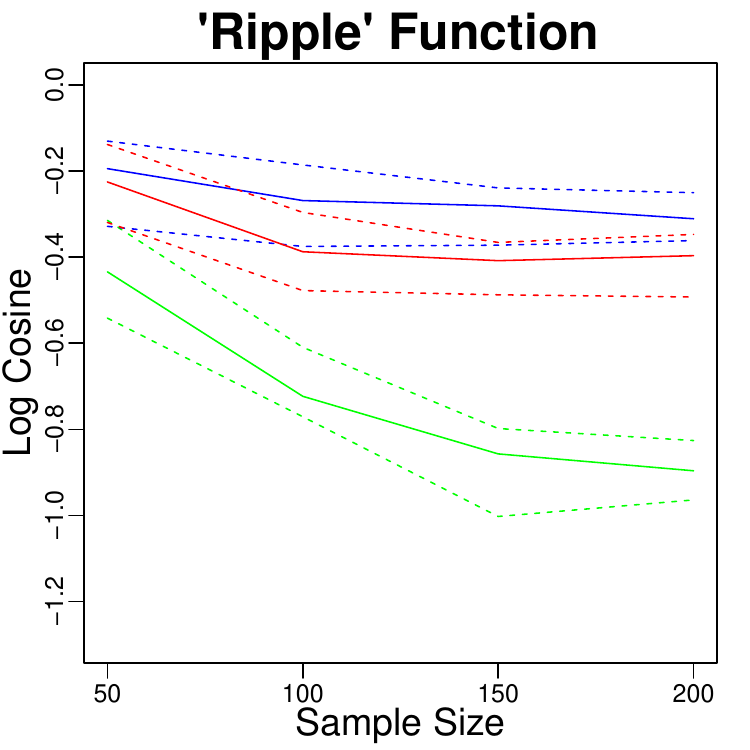}
    
    \caption{\textit{Top:} Visualizations of test functions in 2D. \textit{Bottom:} Mean subspace error for Gaussian Process estimates of the active subspace for various kernel functions.}
    \label{fig:kern}
\end{figure*}

We now study the capability of Gaussian Process surrogates to estimate the active subspace of ridge functions with direction $\ra$ on two smooth functions 
$
f_1(\x) = (\ra^\top\x)^2$
and
$f_2(\x) = e^{-(\ra^\top\x)^2} $
and two discontinuous functions 
$
f_3(\x) = \mathbbm{1}_{[\ra^\top(\x-0.5\mathbf{1})\geq 0]}$ and 
$f_4(\x) = \mathbbm{1}_{[\sin\big(\frac{10\pi}{P}\ra^\top(\x-\frac{\mathbf{1}}{2})\big)\geq 0]}$ (visualized in 2D in Figure \ref{fig:kern}, top).
For each function, with samples sizes $N\in\{50, 100, 150, 200\}$ and in dimensions $P\in\{3,5,7\}$, we fit a Gaussian process with Gaussian, Mat\'ern $\frac{5}{2}$, or Mat\'ern $\frac{3}{2}$ kernels, which lead to infinitely differentiable, twice differentiable, or once differentiable surrogates, respectively \cite[Chapter 4]{williams2006gaussian}.
We compute their active subspace using the \texttt{R} package \texttt{activegp}.
Then, we measure the cosine of the angle between $\ra$ and the leading eigenvector of the estimated active subspace matrix, which serves as our error measure.
We repeat the experiment 30 times, sampling $\ra$ uniformly at random on the unit $P$-sphere.

Figure \ref{fig:kern} shows the results in dimension 7 (the others are qualitatively similar and in the Supplementary Material).
We see that on the smooth functions, the Gaussian and Mat\'ern $\frac{5}{2}$ kernels are better able to exploit smoothness which leads to better subspace estimates.
Conversely, when the function is nonsmooth, the rougher Mat\'ern $\frac{3}{2}$ kernel dominates in terms of error.
Though, strictly speaking, the Mat\'ern covariance is still ``wrong" insofar as the true simulator is not continuous whereas the surrogate is continuously differentiable, it seems that its discontinuous higher order derivatives still allow it to do a better job matching the active subspace than smoother kernels.

%\todo{Since $\Lam^f=\Gam^f$ for sufficiently regular (\todo{phew}) functions, our results about $\Lam^f$ apply directly to existing GP estimates.}

\section{\textit{Flee} Case Study}\label{sec:flee}

\begin{figure*}
    \centering
    \includegraphics[width=0.32\textwidth]{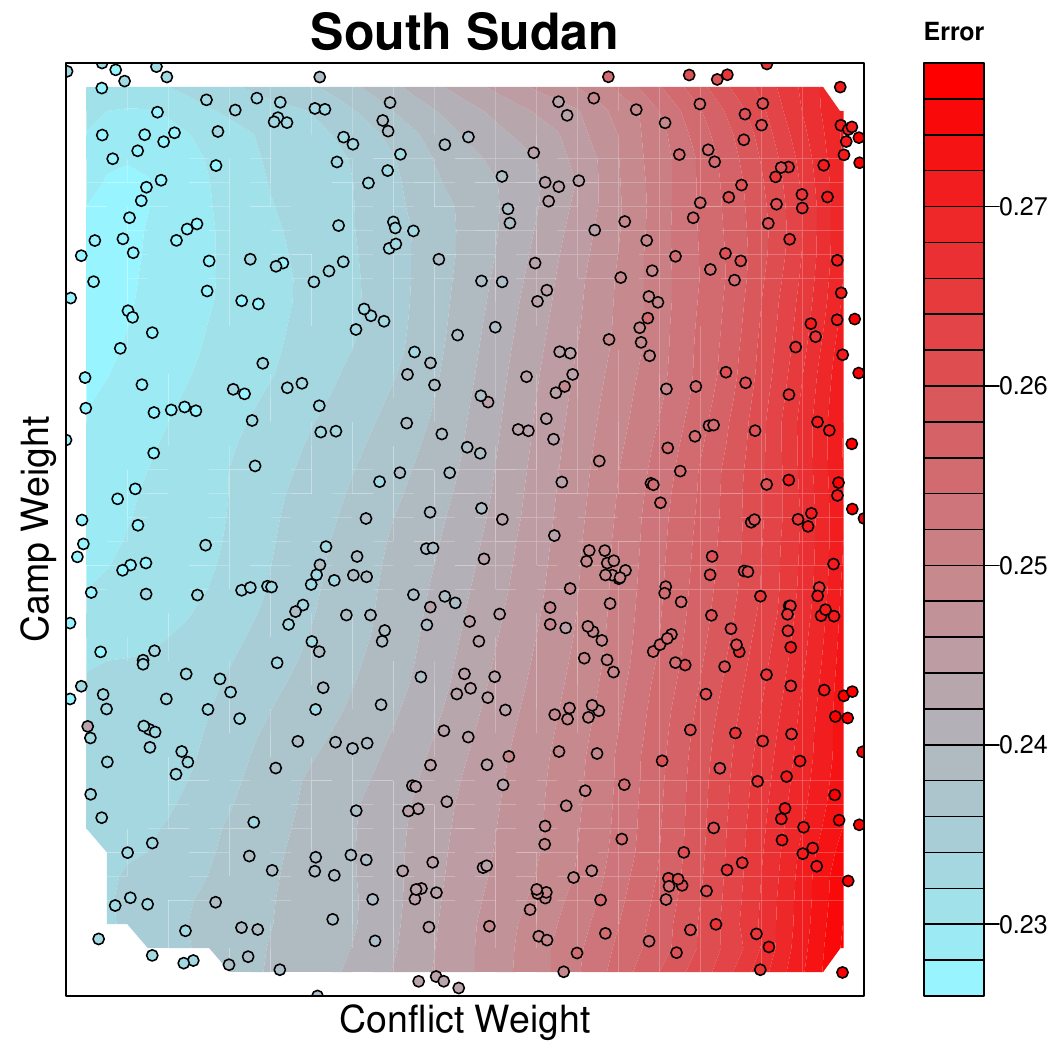}
    \includegraphics[width=0.32\textwidth]{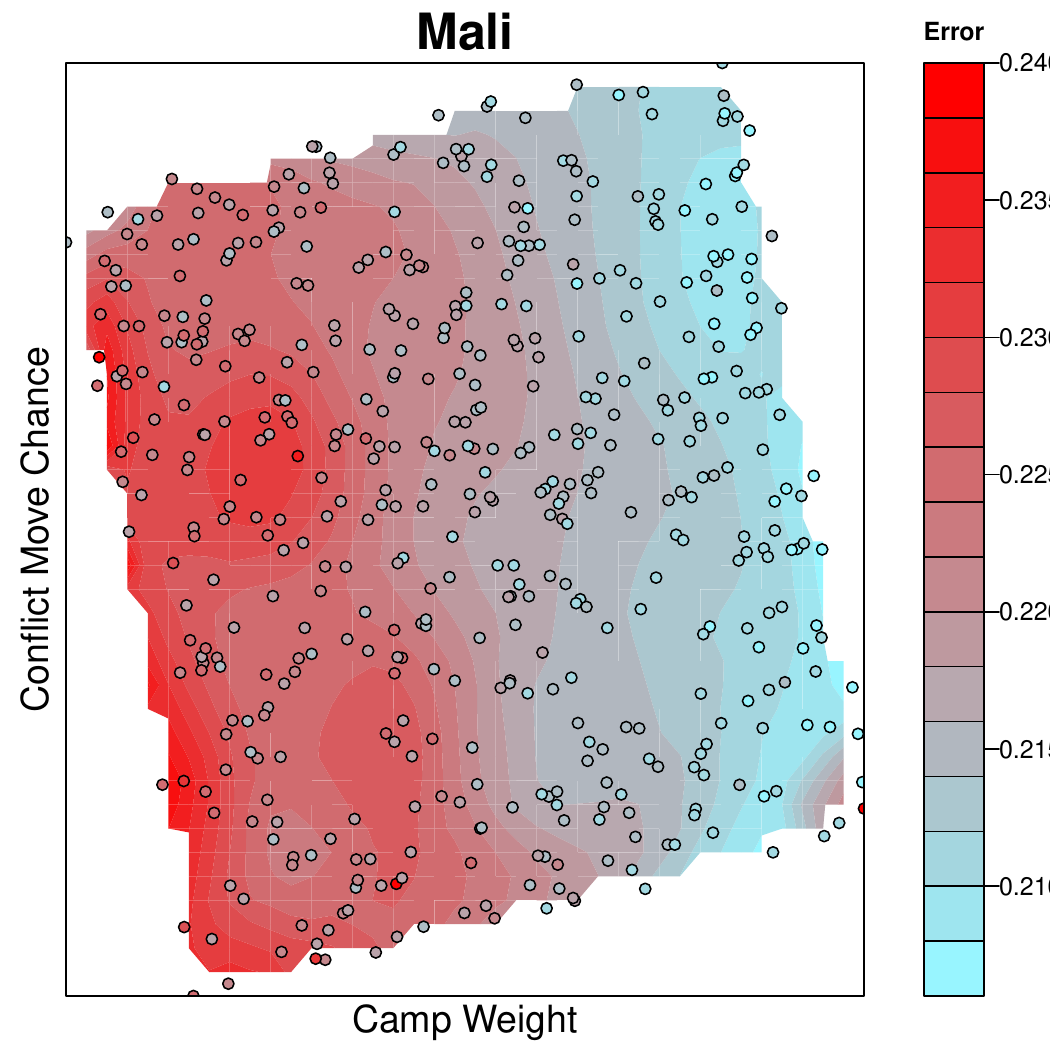}
    \includegraphics[width=0.32\textwidth]{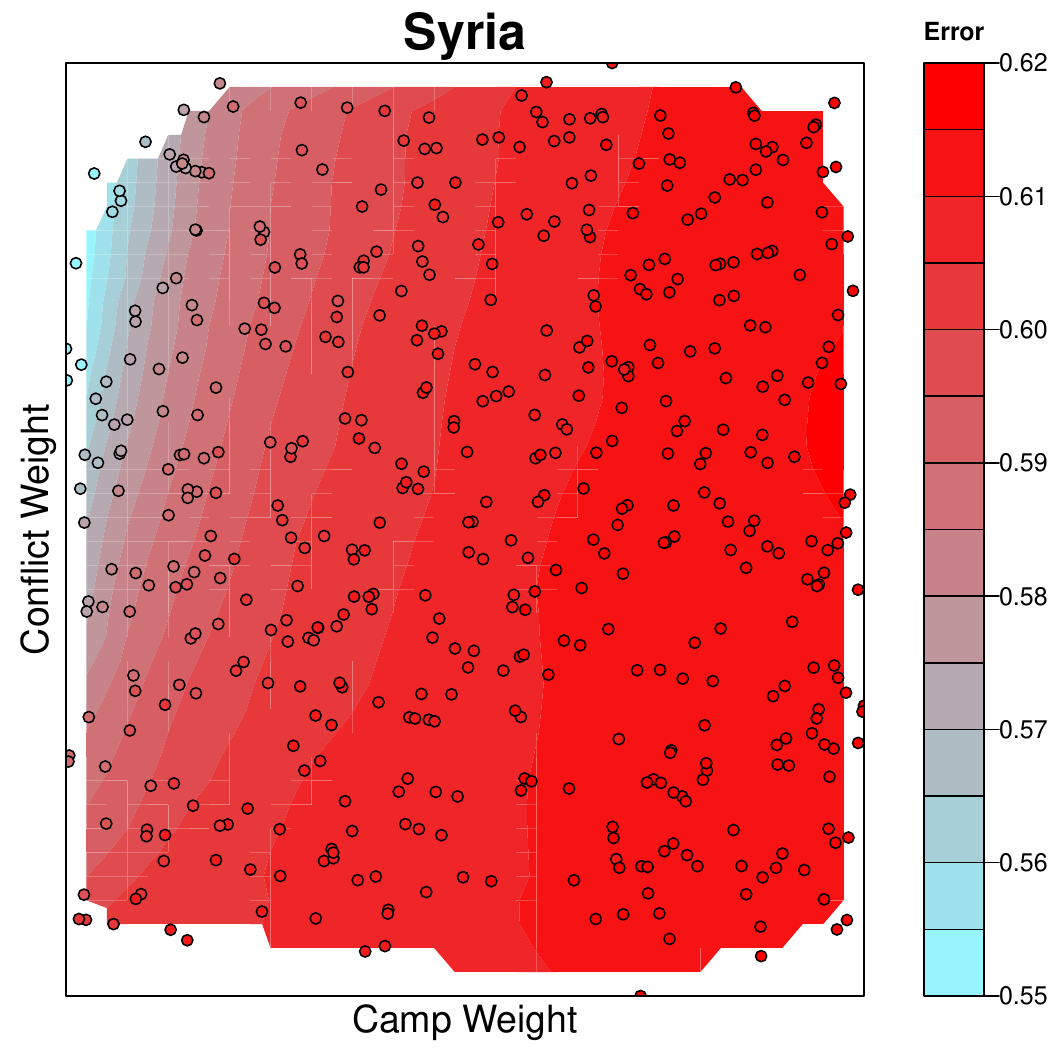}
    
    \includegraphics[width=0.32\textwidth]{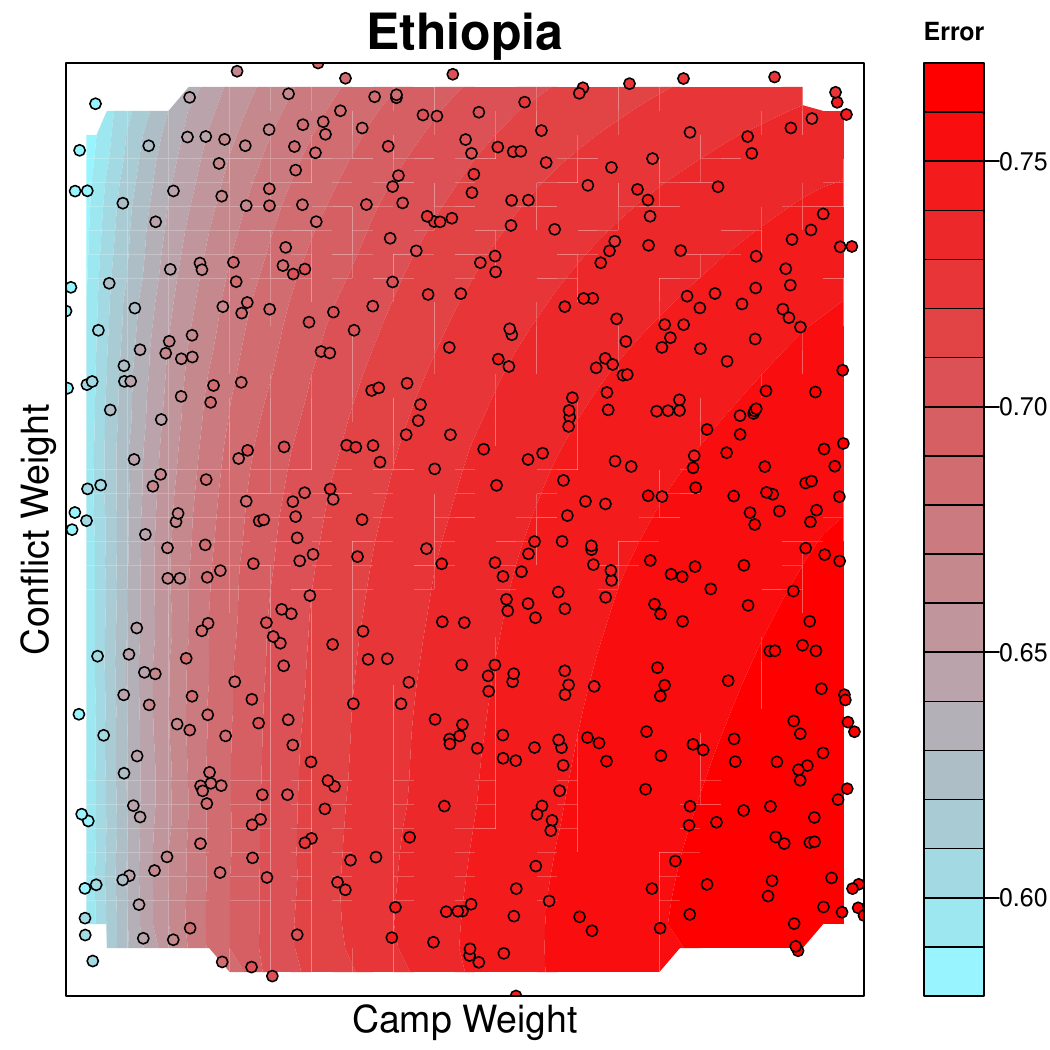}
    \includegraphics[width=0.32\textwidth]{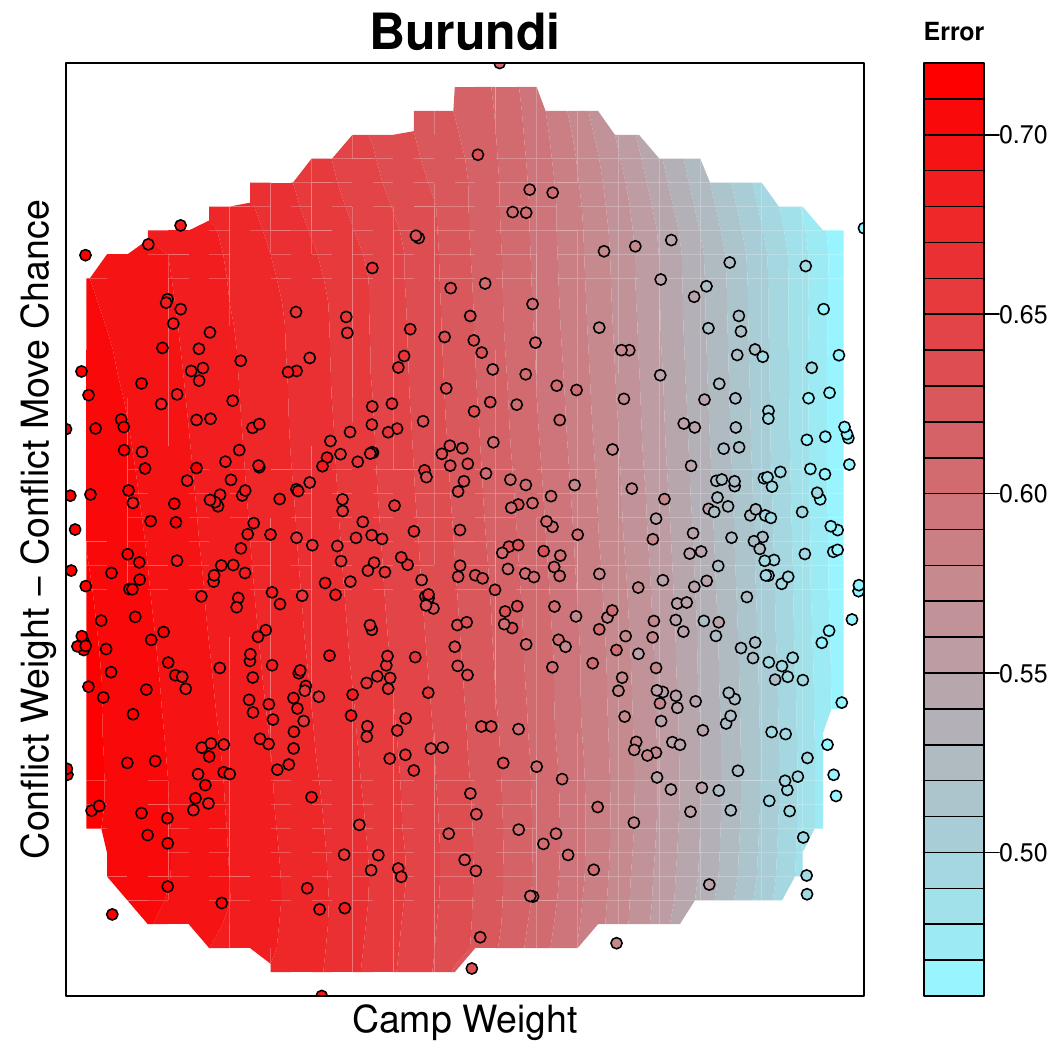}
    \includegraphics[width=0.32\textwidth]{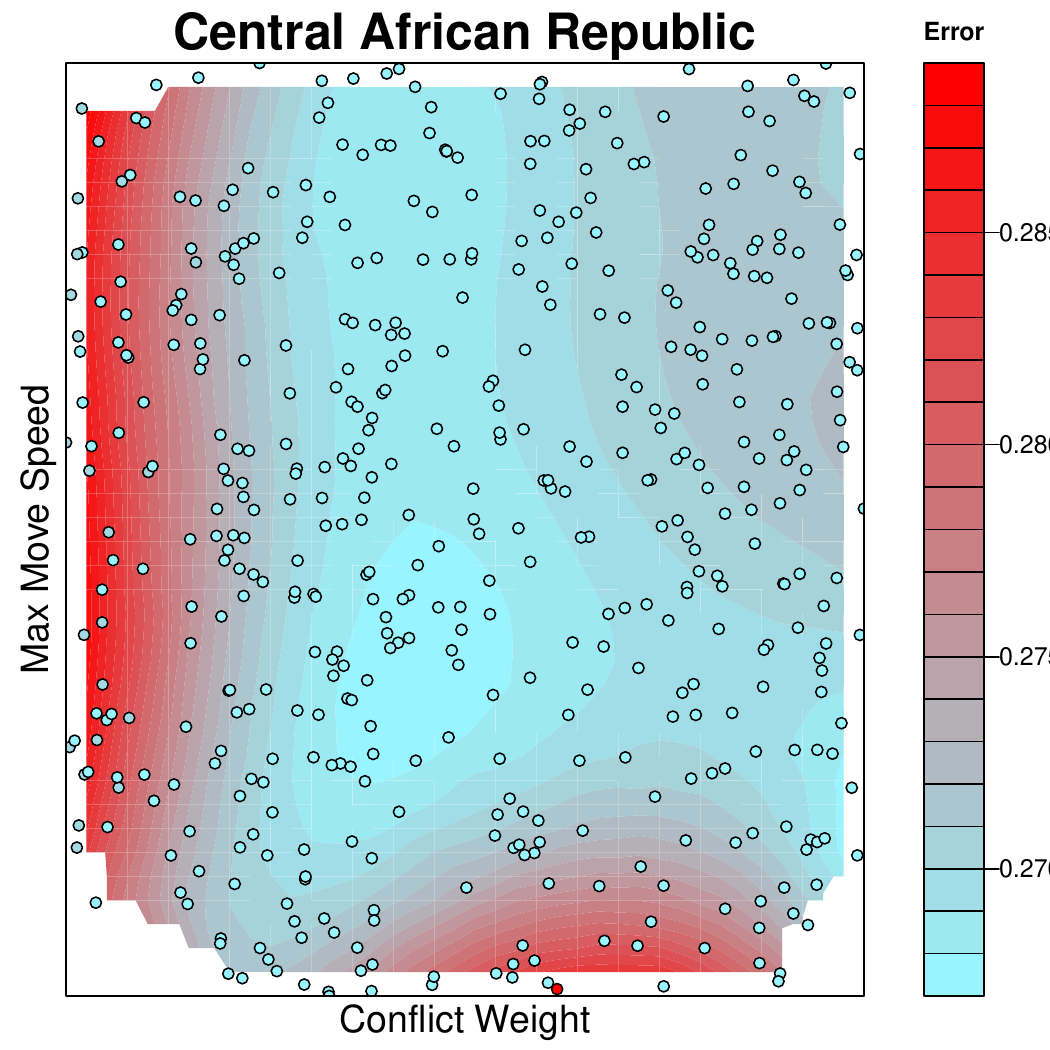}
    
    \vspace{0.5em}
    
    \includegraphics[width=0.95\textwidth]{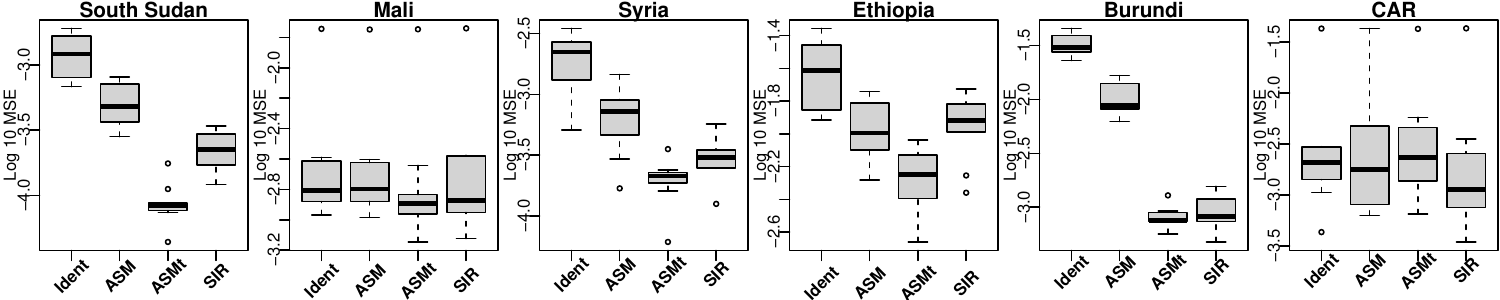}

    \vspace{0.5em}
    
    \includegraphics[width=0.95\textwidth]{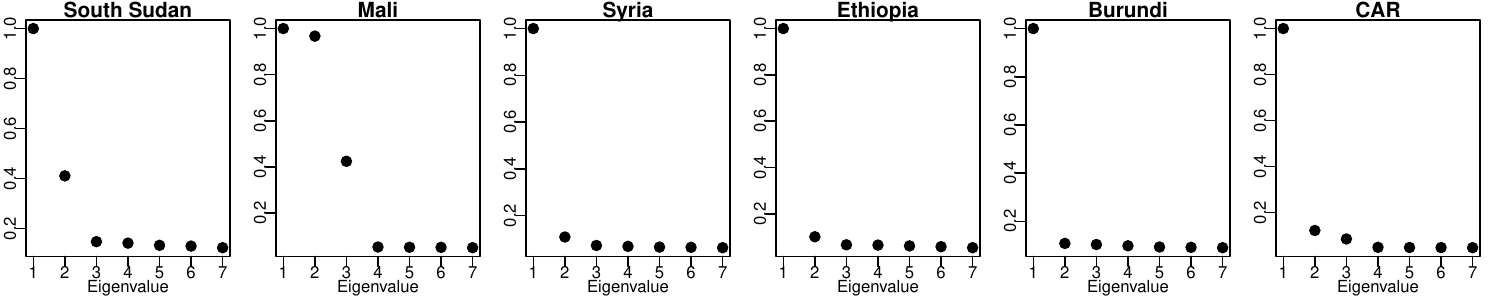}
    
    \caption{Flee simulator case study. \textit{Top:} Active subspace projections of entire data sets.
    \textit{Middle:} 10-fold CV predictive MSE with KNN. Competitors from left to right are KNN with no warping (Ident), with active subspace rotation only (ASM), with active subspace rotation and trucnation (ASMt), and with SIR projection (SIR); lower is better.
    \textit{Bottom:} Eigenvalues of surrogate active subspace matrices for each case study; gaps between subsequent eigenvalues indicate presence of active subspace.
    }
    \label{fig:flee_bakeoff}
\end{figure*}

In this section we deploy active subspaces to the \textit{Flee} ABM (see section \ref{sec:mot_flee}).
We generated a sample of 500 randomly distributed points within the parameter ranges for each of the six case studies and evaluated \textit{Flee} at each of the design points. 
We calculated active subspace estimates using Mat\'ern $\frac{3}{2}$ kernels fit to the entire dataset, and found that the South Sudan study had an active subspace of dimension 2, the Mali study one of dimension 3, and all others one of dimension 1 (Figure \ref{fig:flee_bakeoff}, bottom).

\subsection{Quantitative Prediction Comparison}

To quantitatively evaluate the active subspace sensitivity, we perform K nearest neighbors regression with $K=5$, the default value in the \texttt{caret} R library, optionally ``prewarping" \cite{wycoff2022sensitivity} the points with the active subspace by multiplying them by a square root of the surrogate active subspace matrix.
Furthermore, we truncate the warped data to a lower dimensional space.
We also compare this approach to a projection based on Sliced Inverse Regression, an SDR method.
For truncated active subspace and SIR, we set the dimension of the reduced space to that determined by the eigenanalysis of the active subspace estimated on the full data.
We perform 10-fold CV to estimate predictive accuracy.
We find that the truncated active subspace has lower MSE than the other methods on all datasets except for the Central African Republic case study, which does not show much variation across methods (Figure \ref{fig:flee_bakeoff}, top).
Furthermore, it improves over the original KNN by an order of magnitude on South Sudan, Syria, Ethiopia and Burundi.

\subsection{Qualitative Findings}

Interestingly, we find that the majority of the loadings of the first and second eigenvectors tend to map onto a single variable, with the exception of Burundi's second eigenvector (Table \ref{tab:ev_nums}).
We find that the $\texttt{camp\_weight}$ variable is most important for the Mali, Syria, Ethiopia and Burundi case studies, while $\texttt{conflict\_weight}$ is most important for the CAR and Sudan case studies (Table \ref{tab:ev_qual}).

We compute a projection of the 500 design points using the first two eigenvectors for each case study, shown in Figure \ref{fig:flee_bakeoff}, middle.
For South Sudan, Syria, Ethiopia and Burundi, the surrogate active subspace seems to capture the majority of the variation in the response. 
This is not the case for Mali, which is unsurprising given the fact that the spectrum of the active subspace matrix indicated a three dimensional subspace.
The Central African Republic, on the other hand, has significant outliers not explained by a higher dimensional subspace being present. 
Furthermore, recall that the quantiative study showed little variation across different methods, indicating that linear dimension reduction may not be suitable for this problem.

\begin{table}[ht]
    \tiny
    \centering
    1st Eigenvectors:
    \begin{tabular}{|rrrrrrr|}
      \hline
     Param & S. Sudan & Mali & Syria & Ethiopia & Burundi & CAR \\ 
      \hline
      MMS & 0.01 & -0.00 & -0.00 & -0.00 & -0.00 & -0.00 \\ 
      MWS & -0.01 & 0.00 & 0.00 & 0.00 & -0.00 & -0.00 \\ 
      CMC & 0.00 & -0.00 & 0.00 & -0.01 & 0.00 & -0.00 \\ 
      CoMC & 0.02 & -0.20 & -0.00 & 0.00 & -0.00 & 0.00 \\ 
      DMC & -0.01 & 0.00 & -0.00 & -0.00 & -0.01 & -0.00 \\ 
      CW & 0.01 & 0.98 & 1.00 & 1.00 & 1.00 & 0.00 \\ 
      CoW & 1.00 & 0.00 & -0.01 & 0.03 & -0.04 & 1.00 \\ 
       \hline
    \end{tabular}

    \vspace{1em}

    2nd Eigenvectors:
    \begin{tabular}{|rrrrrrr|}
      \hline
     Param & S. Sudan & Mali & Syria & Ethiopia & Burundi & CAR \\ 
      \hline
      MMS & -0.01 & 0.00 & -0.04 & 0.06 & -0.09 & 1.00 \\ 
      MWS & 0.01 & 0.00 & 0.11 & 0.05 & 0.09 & 0.00 \\ 
      CMC & -0.01 & 0.00 & 0.05 & 0.03 & -0.36 & 0.00 \\ 
      CoMC & 0.02 & 0.98 & -0.00 & 0.01 & -0.61 & -0.00 \\ 
      DMC & 0.01 & 0.00 & -0.06 & -0.03 & 0.33 & -0.01 \\ 
      CW & 1.00 & 0.20 & 0.01 & -0.03 & 0.03 & 0.02 \\ 
      CoW & -0.01 & 0.00 & 0.99 & 1.00 & 0.61 & 0.00 \\ 
       \hline
    \end{tabular}
    \caption{
    Top two eigenvectors of estimated active subspace. 
    MMS is Max Move Speed, MWS is Max Walk Speed, CMC is Camp Move Chance, CoMC is Conflict Move Chance, DMC is Default Move Chance, CW is Camp Weight and CoW is Conflict Weight.
    }
    \label{tab:ev_nums}
\end{table}

\begin{table}[ht]
    \tiny
    \centering
    \begin{tabular}{|rrrrrrr|}
        \hline
         & S. Sudan & Mali & Syria & Ethiopia & Burundi & CAR\\ 
        \hline 
        1 &
        {\color{red} CoW} & 
        {\color{blue} CW} &
        {\color{blue} CW} &
        {\color{blue} CW} &
        {\color{blue} CW} &
        {\color{red} CoW} \\
        2 &
        {\color{blue} CW} & 
        {\color{purple} CoMC} &
        {\color{red} CoW} & 
        {\color{red} CoW} & 
        {\color{red} CoW} - {\color{purple} CMC} &
        {\color{orange} MMS}\\
        \hline
    \end{tabular}
    \caption{Qualitative Representation of Eigenvectors; see Table\ref{tab:ev_nums} caption.}
    \label{tab:ev_qual}
\end{table}

\section{Discussion}\label{sec:disc}

\textbf{Summary:} In this article, we discussed some pathologies associated with surrogate estimation of active subspaces for functions with smooth and discontinuities components, and developed an extension of active subspaces to explain them. 
In our case study, we found that the surrogate active subspace estimates were for the most part axis-aligned.
This was a surprising result; on most case studies to which active subspaces are deployed, the discovered dimensions are combinations of input parameters (e.g. \cite{lukaczyk2014active,constantine2016accelerating,grey2018active}), however, visualization of the projected design points showed that the active subspaces did indeed accurately capture variation in the function, with the exception of the Mali and CAR case studies, which had too high a dimensional subspace or no clear linear subspace, respectively.
Furthermore, it was interesting that both the dimension of the active subspaces and the type of active subspace varied from case study to case study, even for the same simulator and set of parameters.

\textbf{Conclusions:} Our numerical and analytic results provide us with several important conclusions.
In studying a simulator with important smooth and discontinuous structure, we should keep in mind that by choosing a sample size, we are implicitly choosing a tradeoff between them, and that for sufficiently large sample sizes, the smooth directions will be lost.
Furthermore, our analysis, via reasoning by limit arguments, puts into sharp relief a choice that is made when we do active subspaces: by squaring the gradient, we prioritize sharp jumps over gradual ones, even on fully differentiable simulators.
This study also proved the viability of estimating the sensitive directions of a piecewise constant discontinuous function using continuous surrogates, namely Gaussian processes, and our numerical experiments suggest that best accuracy may be achieved by using minimally differentiable kernels, namely the Mat\'ern $\frac{3}{2}$.
In this article we tried to show what analysts are actually estimating when doing surrogate ASM on discrete simulators, in effect cautiously endorsing such analyses.
Another reaction might have been condemnation: why use active subspaces when there are perfectly good dimension reduction tools not reliant on gradients?
Our case study shows that the ASM does better on our application than SIR, a tool which does not use gradient structure, lending an empirical argument for discrete ASM deployment.

\textbf{Future Work:} In revealing some pathologies of the ASM on mixed smooth-discontinuous simulators, we believe we have opened the door to future work which allows for explicit setting of a tradeoff between the two.
One approach would be a hyperparameter governing the relative strength of the two, by decomposing sensitivity into smooth and nonsmooth parts, or by using a different definition which directly avoids the delineated pathologies.
In clarifying the behavior of surrogate active subspaces on fully discontinuous simulators, we hope to lend further theoretical understanding of future applied case studies.
Finally, an implication of our work is that surrogate active subspaces may be useful in the context of mostly smooth simulators with unknown discontinuities.
Whereas \cite{gorodetsky2014efficient} develop an algorithm for determining \textit{where} discontinuities occur, this proposed future work would determine along which \textit{directions} discontinuities occur simply by conducting surrogate active subspace analysis on the simulator with a sufficiently large sample size.
Some work would be required to determine how large is large enough, and how to best benefit from knowledge of these directions.

%\begin{enumerate}
%    %\item May be useful as an ``edge detector" in discontinuous surrogate modeling (since nonsmooth overwhelms smooth part). Whereas \cite{gorodetsky2014efficient} develop an algorithm for determining \textit{where} discontinuities occur, this proposed future work would determine along which \textit{directions} discontinuities occur.
%    %\item Otherwise, may want to avoid if we care about both smooth and discontinuous structure: even if we only care about one, by choosing a sample size, we are implicitly choosing a tradeoff.
%    %\item This analysis, via reasoning by limit arguments, also puts into sharp relief a choice that is made when we do active subspaces: to prioritize sharp jumps over smooth ones. 
%    %\item If using a GP estimate for , may want to use as sharp a covariance as able.
%    %\item In this article we tried to show what people are doing when they do surrogate ASM on discrete simulators. Another reaction might be condemnation: why use active subspaces when there are perfectly good dimension reduction tools not reliant on gradients? Our case study shows that the ASM does better on our application than SIR, a tool which does not use gradient structure, lending an empirical argument for discrete ASM deployment.
%\end{enumerate}

%\subsubsection*{Acknowledgements}
%Anonymized for review.

\bibliography{main}
\bibliographystyle{apalike}

\end{document}